\documentclass{article}

\usepackage[final]{corl_2026} 
\newcommand{\method}{\textsc{ProbeAct}\xspace}
\usepackage{xspace}

\usepackage{booktabs} 
\usepackage{multirow}
\usepackage{amsmath}
\usepackage{wrapfig}
\usepackage{graphicx}
\usepackage{enumitem}

\usepackage{amssymb}

\title{ProbeAct: Probe-Guided Training-Free Failure Recovery in Vision-Language-Action Models}

\author{
  Fan Zhang, Seongbin Park, Baharan Mirzasoleiman, Shahriar Talebi, Nader Sehatbakhsh\\
  University of California Los Angeles 
  United States\\
  \texttt{parkseongbin@ucla.edu} \\
}

\begin{document}
\maketitle


\begin{abstract}
    Vision-Language-Action (VLA) models demonstrate strong performance on language-conditioned robotic manipulation within their training distribution, yet their generalization capabilities remain fundamentally limited. They lack the robustness required to handle  perturbations, frequently failing when confronted with lighting changes, altered camera viewpoints, or small initial-state variations. We propose \method, a training-free runtime intervention framework that detects and recovers from grasping and placement failures in pretrained VLA policies without modifying their weights or requiring additional demonstrations. \method combines three components: (i) a lightweight \emph{multi-target hidden-state probe} that predicts the 3D positions of task-relevant objects from intermediate VLA features, with Hungarian-matched identity tracking for multi-object scenes; (ii) an \emph{object-agnostic kinematic state machine} that detects grasp, transport, and placement failures using only gripper-internal signals and end-effector kinematics; and (iii) a \emph{hierarchical Control Barrier Function (CBF) filter} that encodes repeated-failure locations as soft safe-set constraints, minimally correcting VLA actions while preserving baseline behavior. As a plug-and-play, training-free intervention loop, \method is orthogonal to existing training pipelines. Evaluated on the LIBERO-plus benchmark, our framework acts as a universal safety net, improving the success rate of the OpenVLA-OFT model from 69.6\% to 74.1\%, while demonstrating broad applicability to both base and fine-tuned VLA policies.
\end{abstract}



\section{Introduction}
 
Vision-Language-Action (VLA) models~\cite{kim2024openvla, kim2025fine, black2024pi_0, intelligence2025pi_}, predominantly trained via behavioral cloning on large-scale expert demonstrations, have emerged as the dominant paradigm for language-conditioned robotic manipulation. Despite achieving impressive proficiency within their training distributions, these models exhibit profound brittleness when subjected to realistic out-of-distribution (OOD) perturbations. Empirical evaluations demonstrate that even minor environmental shifts, such as altered lighting, modified textures, visual distractors, or slight deviations in initial object configurations, can cause success rates to drop precipitously~\cite{fei2025libero, zhou2025libero}. Recent analyses~\cite{goel2025geometric, zeng2025diagnose, yang2026uaor} converge on a critical diagnosis for this fragility: constrained by the behavioral cloning objective, VLAs tend to overfit to their nominal training trajectories. Consequently, when test scenes diverge from the training conditions, the policy fails to dynamically ground its actions in the current visual context. Instead, it bypasses active spatial reasoning and blindly reproduces a memorized kinematic sequence.
 
Critically, this memory trap manifests as a stark decoupling between latent perception and motor execution~\cite{zhou2025libero,fang2026vision, sendai2025leave}. As corroborated by our own probing experiments, the VLA's visual backbone successfully processes the perturbed scene, maintaining accurate spatial representations of the target object. The failure bottleneck lies exclusively within the action head: constrained by an overfitted mapping, the network collapses to the nominal trajectory of its training distribution. We systematically observe this precise failure mode when evaluating OpenVLA-OFT on the LIBERO-plus benchmark~\cite{fei2025libero}. 
 
A complementary line of work mitigates OOD failures at \emph{inference time} via online corrections~\cite{zeng2025diagnose, shah2025learning, lin2025failsafe, ma2026cyclevla}. However, prevailing approaches introduce significant structural overhead by relying on external reasoning modules, additional visual symbols, or explicit 3D reprojection. They often demand supplementary sensing infrastructure or external VLMs to completely override the VLA's native action stream. In contrast, \method demonstrates that the geometric reference can be extracted directly from the VLA's internal hidden states. Since our analysis establishes that the VLA maintains accurate spatial features despite action head collapse, external sensors are redundant. We frame the recovery as a minimal spatial constraint, safely redirecting the policy away from the memory trap while maximally preserving its original kinematic intent.


We operationalize this via \method, a training-free intervention framework comprising three coupled components:
(1) Perception via Hidden-State Probing. We train a multi-target probe with online Hungarian matching~\cite{kuhn1955hungarian} to extract stable 3D coordinates directly from intermediate VLA activations, eliminating the need for external vision models. (2) Object-Agnostic Failure Detection. Rather than using task-specific supervision, we reliably detect execution anomalies (e.g., empty grasps, drops) by evaluating \emph{relative state changes} and \emph{kinematic synchronization} between the robot proprioception and probe-grounded targets. (3) Hierarchical CBF Safety Filtering. We formulate repeated failures as a runtime safety problem. Escaping catastrophic failure loops, \method escalates from a stateless kinematic recovery to a stateful Control Barrier Function (CBF)~\cite{ames2019control} constraint, resolving the memory trap via minimal closed-form action projection while preserving baseline competence.

\textbf{Summary of Contributions.}  
\textbf{(1)} We propose \method, a lightweight, closed-loop intervention framework that extracts robust 3D geometric references directly from frozen VLA hidden states, entirely eliminating the structural overhead of external visual sensors. 
\textbf{(2)} We design an object-agnostic kinematic state machine coupled with a hierarchical Control Barrier Function (CBF) filter. This translates relative spatial synchronization into a closed-form action projection, autonomously rescuing the policy from catastrophic failure loops while preserving baseline competence. 
\textbf{(3)} Extensive evaluation on the LIBERO-plus benchmark demonstrates consistent success rate improvements across diverse visual and spatial perturbation categories, achieved entirely at inference time without VLA weight modification, external reasoning modules, or additional demonstrations.

\section{Related Work}

\textbf{Vision-Language-Action Models.}
OpenVLA~\cite{kim2024openvla} fine-tunes a LLaMA-7B backbone on the Open X-Embodiment dataset to directly map image and language inputs to robot actions. OpenVLA-OFT~\cite{kim2025fine} extends this paradigm via one-shot fine-tuning and an L1 regression action head, significantly improving sample efficiency. Concurrently, systems such as RT-2~\cite{zitkovich2023rt}, $\pi_0$~\cite{black2024pi_0}, and Octo~\cite{team2024octo} have explored alternative architectures and training objectives for generalist manipulation. Despite these foundational advances, modern VLAs remain highly susceptible to catastrophic failures under out-of-distribution (OOD) shifts. This fragility motivates our focus on \emph{training-free runtime recovery} rather than solely pursuing improved pre-training methodologies.

\textbf{Probing Neural Policies.}
Linear probing has been extensively utilized to extract structured semantic information from the hidden states of deep neural networks~\cite{alain2016understanding, belinkov2022probing, lu2025probing}. In the context of robotics, prior studies have employed probes on pre-trained visual and VLA features to extract spatial geometry~\cite{el2024probing} and symbolic object or action states. However, these techniques have primarily served as retrospective analytical tools to understand representation hierarchies. \method departs from this paradigm by repurposing probe outputs as active, runtime sensor signals to drive closed-loop intervention decisions. 

\textbf{Control Barrier Functions in Robotics.}
Control Barrier Functions (CBFs) provide a rigorous mathematical framework for enforcing safety constraints in dynamical systems~\cite{ames2016control, marley2021maneuvering}. While CBFs have been widely adopted in autonomous driving~\cite{kim2024control}, legged locomotion~\cite{ames2016control}, and collision avoidance~\cite{shen2021collision}, our integration of CBFs into VLA policies is novel in two key aspects: (i) the safe set is constructed \emph{dynamically online} based on observed kinematic failures rather than being specified \emph{a priori}; and (ii) the system formulates a \emph{minimal correction} to the pre-trained policy's action sequence, bridging classical CBF safety guarantees with the modern paradigm of large, end-to-end foundation models.

\textbf{Failure Recovery in Manipulation.}
Traditional failure recovery in robotic manipulation has heavily relied on hierarchical task and motion planning (TAMP)~\cite{pan2022failure}, reactive replanning heuristics using behavior trees~\cite{colledanchise2016behavior}, or explicitly learned recovery policies via reinforcement learning~\cite{thananjeyan2021recovery}. These conventional approaches typically demand rigorous environment modeling, known failure probabilities, or prohibitively expensive failure-case training data. To our knowledge, \method is the first framework to synthesize probe-based latent perception with CBF-constrained kinematic correction, achieving fully \emph{training-free} and \emph{object-agnostic} failure recovery for pre-trained VLAs.
\section{Method}
\label{sec:method}

\begin{figure}
    \centering
    \includegraphics[width=0.97\linewidth]{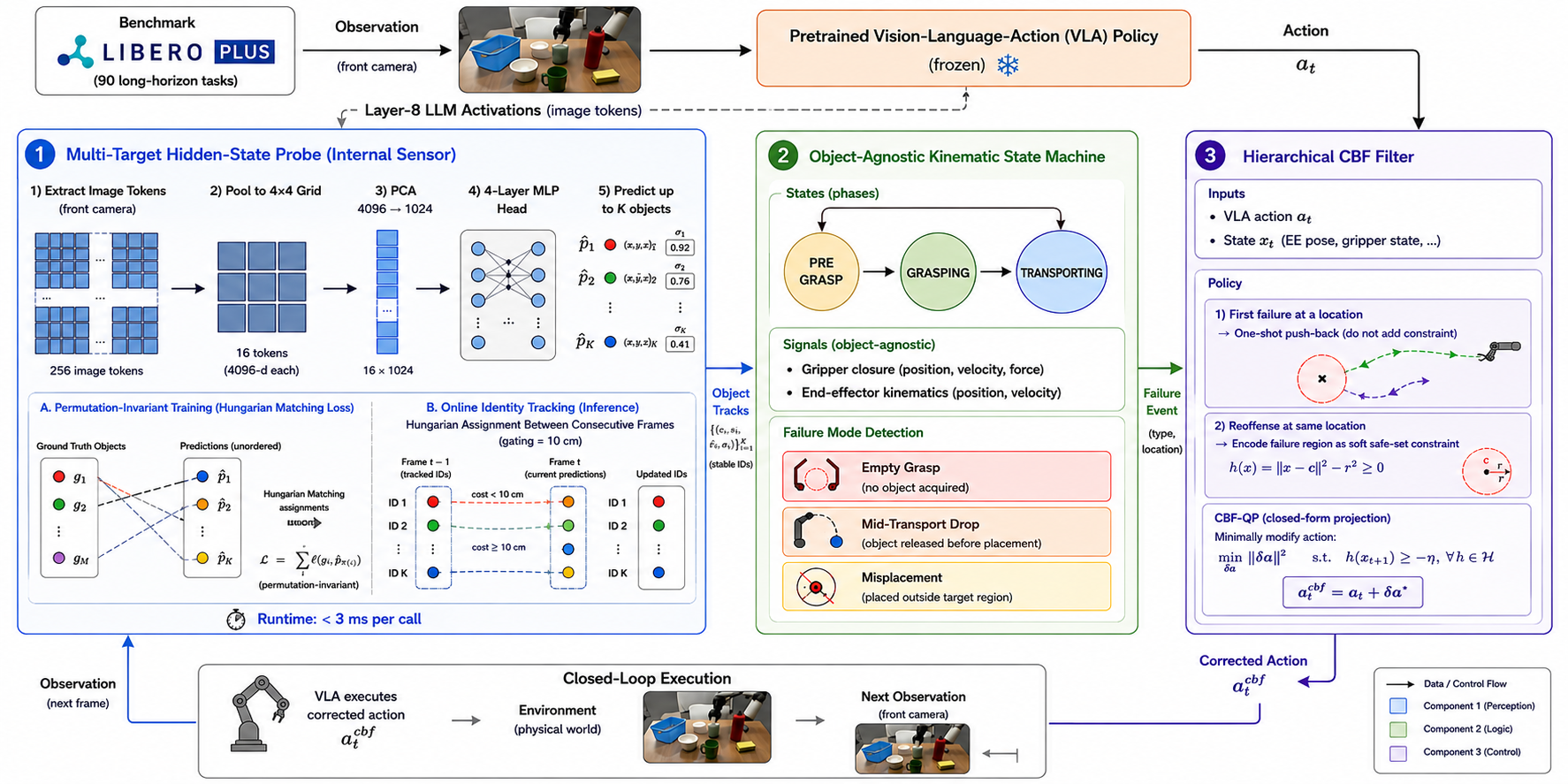}
    \caption{\textbf{Overview of the \method Framework.} Operating entirely at inference time alongside a frozen VLA, the system consists of three modules: (1) an internal probe that extracts stable 3D object tracks from intermediate LLM activations; (2) a kinematic state machine that detects physical execution failures via relative object-robot synchronization; and (3) a hierarchical CBF filter that minimally modifies the nominal action $a_t$ into a safe action $a_t^{cbf}$ to escape failure zones. The framework forms a closed-loop recovery mechanism without requiring external 3D sensors.}
    \label{fig:method_overview}
    \vspace{-5pt}
\end{figure}

\subsection{Overview}
 
We propose \method, a training-free, inference-time intervention framework that rescues frozen Vision-Language-Action (VLA) models from memory traps (Figure~\ref{fig:method_overview}). Bypassing the need for external 3D sensors, \method operates continuously via three streamlined components: \textbf{(1) Perception:} An internal probe extracts stable 3D object coordinates directly from intermediate VLA activations. \textbf{(2) Logic:} A kinematic state machine detects physical execution failures (e.g., empty grasps) by evaluating relative robot-object synchronization. \textbf{(3) Control:} Upon detection, a hierarchical Control Barrier Function (CBF) filter applies a minimal closed-form projection to the nominal action $a_t$, safely deflecting the trajectory away from the failure zone without disrupting baseline competence.
 
\subsection{Multi-Target Hidden-State Probe}
\label{sec:probe}
 
\textbf{Architecture.} We train a multi-head MLP $\phi:\mathbb{R}^{d_\text{pca}} \to \mathbb{R}^{K\times 3} \times \mathbb{R}^K$ that maps a PCA-reduced spatial feature (from VLA layer 8) to up to $K$ object positions with confidence scores. The feature is obtained by averaging the 16$\times$16 spatial tokens into a 4$\times$4 grid of 4096-dim vectors, then applying PCA to obtain $d_\text{pca}{=}1024$ dimensions. Each predicted position $\hat{p}_k \in \mathbb{R}^3$ is accompanied by a confidence score $c_k \in [0,1]$ obtained via a sigmoid activation. Predictions with $c_k < 0.5$ are safely discarded. The probe is parameterized as a 4-layer MLP with hidden dimensions $[2048, 1024, 512, 256]$. Each hidden layer is sequentially followed by Layer Normalization, a ReLU non-linearity, and dropout ($p=0.1$).
 
\textbf{Training with Hungarian matching.} Frames contain variable numbers of relevant objects (1 to $K$), so we optimize the probe using a bipartite matching loss. Let $M^{(n)} = |p^{(n)}|$ be the number of ground-truth objects in frame $n$. We pad the ground truth with $\varnothing$ (no object) up to $K$ slots and define a target mask $m_i \in \{0,1\}$, where $m_i=1$ if $i \le M^{(n)}$ and $0$ otherwise. The Hungarian matching loss is defined as:
\begin{equation}
\mathcal{L} = \frac{1}{N}\sum_{n=1}^{N} \min_{\sigma \in \mathfrak{S}_K} \left[ \sum_{i=1}^{M^{(n)}} \|\hat{p}_{\sigma(i)}^{(n)} - p_i^{(n)}\|_2^2 \;+\; \beta \sum_{i=1}^K \text{BCE}(c_{\sigma(i)}^{(n)}, m_i) \right]
\label{eq:hungarian}
\end{equation}
where $\mathfrak{S}_K$ denotes the symmetric group of all permutations of $K$ elements. The optimal permutation $\sigma$ matches predicted slots to ground-truth objects to minimize the joint localization and confidence error. Consequently, the probe's output is \emph{permutation-invariant}: there is no enforced temporal consistency in the slot indices, meaning slot $k$ at frame $t$ does not inherently correspond to slot $k$ at frame $t+1$.
 
\textbf{Inference-time identity tracking.}
While permutation invariance facilitates training, the lack of temporal consistency is problematic for downstream closed-loop control; the system must maintain a persistent identity for each physical object across consecutive frames. We resolve this via \emph{online Hungarian matching} over the temporal axis. At each timestep $t$, newly predicted coordinates are assigned to existing object tracks by minimizing the pairwise Euclidean displacement. To prevent erroneous ID switching from noise or false positives, we apply a spatial gating threshold bounded by the maximum plausible inter-frame object velocity. Unmatched predictions beyond this boundary automatically spawn new tracks. This lightweight procedure yields the temporally stable 3D position estimates required for the multi-step failure detection logic detailed in \S\ref{sec:state_machine}.
 
\subsection{Object-Agnostic Kinematic State Machine}
\label{sec:state_machine}

\begin{figure}[t]
    \centering
    \includegraphics[width=0.99\linewidth]{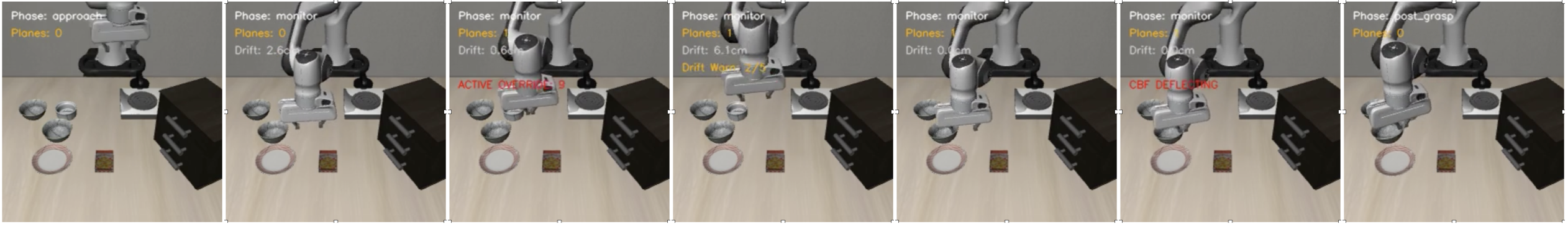}
    \caption{\textbf{Real-Time Failure Detection and Intervention.} A sequential rollout demonstrating the state machine in action. As the VLA policy diverges from the true target (exhibiting spatial drift during the monitoring phase), the kinematic state machine detects the decoupling and triggers an active override. The hierarchical CBF filter immediately deflects the end-effector trajectory, safely recovering the grasp.}
    \label{fig:process}
\end{figure}
\vspace{-5pt}

We design failure detection around six mechanical phases: {\sc Approach}, {\sc Monitor}, {\sc Grasping}, {\sc Post grasp} and one event, {\sc Placed}. All transitions are evaluated using \emph{object-agnostic} signals: gripper width $q$, end-effector pose $\mathbf{e}=(x_e, y_e, z_e)$, and tracked object positions from the probe. To ensure generalizability without sacrificing mathematical rigor, the state machine avoids task-specific hyperparameter tuning. Instead, it evaluates relative state changes and kinematic synchronization, thresholded purely by embodiment-specific hardware tolerances (e.g., nominal lift clearance).

\textbf{Pre Grasp $\to$ Grasping.} 
The system transitions to the {\sc Grasping} phase when the end-effector approaches a probe-tracked target within a specific distance ($\|\mathbf{e} - p_\text{obj}\| \le \rho_\text{enter}$). At this moment, we record the initial grasping state: the target's identity, its starting position $p_\text{enter}$, and the starting end-effector height $z_\text{enter}$. These recorded values act as the baseline for calculating all subsequent relative movements.

\textbf{Grasp Verification via Kinematic Synchronization.}
A successful grasp is inherently characterized by the synchronous movement of the robotic end-effector and the target object. Over a rolling observation window, we calculate the relative vertical displacements: $\Delta z_e = z_e - z_\text{enter}$ and $\Delta z_\text{obj} = z_\text{obj} - z_{\text{obj,enter}}$. A grasp is confirmed only when the gripper stabilizes above its mechanical minimum ($q > \epsilon_\text{limit}$) and both entities exhibit \emph{synchronous positive displacement} beyond the hardware noise floor ($\Delta z_e > \tau_\text{lift}$ and $\Delta z_\text{obj} > \tau_\text{track}$). By checking that the object actually rose, we avoid marking empty-air grasps as successful.

\textbf{Grasping Failure Modes.}
We detect two distinct failure conditions during this phase, both of which trigger immediate intervention. \textbf{(1) Hard empty grasp:} The gripper width converges to its physical minimum ($q \le \epsilon_\text{limit}$), definitively indicating a complete closure with no object between the fingers. \textbf{(2) Soft empty grasp:} The end-effector exhibits significant upward displacement ($\Delta z_e > \tau_\text{lift}$) while the target object remains stationary ($\Delta z_\text{obj} \le \tau_\text{noise}$). This kinematic decoupling provides definitive evidence that the object was not acquired.

\textbf{Mid-transport drop.} 
During {\sc Post Grasp}, a dropped object is detected through an abrupt kinematic collapse: the gripper width suddenly snaps to its mechanical limit ($q \le \epsilon_\text{limit}$) while the end-effector is actively in motion. The drop location is subsequently encoded as a CBF zone (\S\ref{sec:cbf}), and the system triggers a recovery action to the original grasp target.

\textbf{Placement verification.}
Placement verification relies on a strict sequence of relative state changes to reject premature releases. We trigger {\sc Placed} only when three conditions hold simultaneously: (a) the gripper width increases relative to its holding state ($q > q_\text{hold} + \epsilon_\text{release}$); (b) the end-effector establishes a diverging retreat trajectory ($\|\mathbf{e}_t - p_\text{obj}\| > \|\mathbf{e}_{t-1} - p_\text{obj}\|$); and (c) the held object's spatial coordinates stabilize within the tolerance of the target location.
 
\subsection{Hierarchical Control Barrier Function Filter}
\label{sec:cbf}

Adding a permanent safety zone for every minor error overly restricts the robot and harms its baseline success rate. To avoid this, \method employs a two-tier hierarchical strategy. For an initial failure, the system simply executes a stateless push-back, allowing the VLA to freely attempt self-correction. However, if the robot fails again in the same spatial region, we diagnose a memory trap. Only then do we instantiate a persistent Control Barrier Function (CBF) zone at the failure coordinate $c$ with safety radius $r_\text{safe}$. The barrier function is $h(x; c, r_\text{safe}) = \|x - c\|^2 - r_\text{safe}^2$, defining the safe set $\mathcal{H} = \{x : h(x) \geq 0\}$.

During the {\sc Pre\_Grasp} phase, if any safety zones are active, the VLA's nominal translational action $u_\text{vla} \in \mathbb{R}^3$ is continuously filtered to satisfy the first-order CBF condition $\nabla h(x)^\top u \geq -\gamma h(x)$. This is solved via a minimal-intervention Quadratic Program, yielding the closed-form correction:
\begin{equation}
u_\text{filtered} = u_\text{vla} + \max\!\left(0,\;\frac{-\gamma h(x) - \nabla h(x)^\top u_\text{vla}}{\|\nabla h(x)\|_2^2}\right) \nabla h(x),
\label{eq:cbf_filter}
\end{equation}
where $\nabla h(x) = 2(x - c)$. If the VLA's proposed action naturally respects the safety boundary, the geometric projection evaluates to zero, rendering the filter an exact identity mapping that preserves baseline competence. To prevent the artificial accumulation of constraints across tasks, zones are dynamically flushed when the end-effector enters the terminal proximity ($\rho_\text{clear}$) of the true target or upon triggering the {\sc Placed} event. Furthermore, the filter is intentionally bypassed during the {\sc Grasping} and {\sc Transporting} phases to allow unhindered physical interaction with the object.
 
\subsection{Multi-Step Task Support}
\label{sec:multistep}
 
For tasks involving multiple pick-place pairs (e.g., \emph{``put both moka pots on the stove''}), we parse the task language for keywords (both, two, and) and decompose into $N$ (pick, place) subtask pairs. After {\sc Placed}, the system: (a) appends the just-placed object to a blacklist used by target matching, (b) resets the probe tracker, (c) clears all CBF zones (the next pick begins fresh). The kinematic state machine then re-enters {\sc Pre\_Grasp} for the next pick. This design isolates failure modes across subtasks while reusing the same detection infrastructure.

\section{Experiments}
 
\textbf{VLA backbone.} We use OpenVLA-OFT~\cite{kim2025fine} as our baseline policy. 
\textbf{Benchmark.} We evaluate on LIBERO-plus~\cite{fei2025libero}, a  benchmark covering seven perturbation categories: \emph{Background Textures}, \emph{Lighting condition}, \emph{Camera Viewpoints}, \emph{Robot Initial States} , \emph{Background Textures}, \emph{Sensor Noise} and \emph{Language Instructions}.  Success is measured by LIBERO's built-in goal predicate.
\textbf{Probe training.} We collect ~50{,}000 (hidden-state, object-positions) pairs by rolling out the baseline VLA. Object positions are extracted from the simulator's 
\texttt{obj\_of\_interest} list. The probe trains for 200 epochs (AdamW, batch 512, cosine LR schedule) using the Hungarian loss in Eq.~\ref{eq:hungarian}. 
\textbf{Compute.} All experiments use two NVIDIA RTX PRO 6000 (Blackwell) GPUs. 
 
\subsection{Main Results}

Table~\ref{tab:performance_comparison} compares \method against the baseline OpenVLA-OFT and seven other state-of-the-art VLAs across the seven LIBERO-plus categories. \method achieves the highest overall success rate (\textbf{74.1\%}), demonstrating that robust inference-time intervention can surpass expensive retraining paradigms. More importantly, the distribution of these improvements strongly validates our core hypothesis regarding VLA out-of-distribution (OOD) failures.

\begin{table}[tbp]
\centering
\caption{Success rates (\%) across LIBERO-plus perturbation categories. \method achieves the highest overall score among all evaluated VLAs.}
\label{tab:performance_comparison}
\resizebox{\textwidth}{!}{%
\begin{tabular}{lcccccccc}
\toprule
Model & Camera & Robot & Language & Light & Background & Noise & Layout & Total \\
\midrule
OpenVLA        & 0.8  & 3.5  & 23.0 & 8.1  & 34.8 & 15.2 & 28.5 & 15.6 \\
NORA           & 2.2  & 37.0 & 65.1 & 45.7 & 58.6 & 12.8 & 62.1 & 39.0 \\
WorldVLA       & 0.1  & 27.9 & 41.6 & 43.7 & 17.1 & 10.9 & 38.0 & 25.0 \\
UniVLA         & 1.8  & 46.2 & 69.6 & 69.0 & 81.0 & 21.2 & 31.9 & 43.9 \\
$\pi_0$        & 13.8 & 6.0  & 58.8 & 85.0 & 81.4 & 79.0 & 68.9 & 53.6 \\
$\pi_0$-Fast   & 65.1 & 21.6 & 61.0 & 73.2 & 73.2 & 74.4 & 68.8 & 61.6 \\
RIPT-VLA       & 55.2 & 31.2 & 77.6 & 88.4 & 91.6 & 73.5 & 74.2 & 68.4 \\
\midrule
OpenVLA-OFT    & 56.4 & 31.9 & 79.5 & 88.7 & 93.3 & 75.8 & 74.2 & 69.6 \\
\textbf{\method} (\textit{ours}) & \textbf{63.8} & \textbf{40.3} & \textbf{82.0} & \textbf{93.6} & \textbf{93.5} & \textbf{76.8} & \textbf{80.9} & \textbf{74.1} \\
\bottomrule
\end{tabular}
}
\end{table}
\vspace{-5pt}

\textbf{Spatial Shifts Trigger Memory Traps.} 
The most substantial performance amplifications emerge under geometric distribution shifts, specifically \emph{Robot Initial States} and \emph{Camera Viewpoints}. Under these perturbations, the baseline policy typically exhibits the exact ``Phantom Grasp'' pathology defined in Section~\ref{sec:method}: it identifies the correct target but its motor execution spatially collapses to a memorized training mean. Because \method explicitly evaluates relative kinematic synchronization rather than absolute coordinates, it perfectly intercepts these spatial offsets, allowing the CBF filter to deflect the trajectory back to the true target.

\subsection{Generalization to Fine-tuned VLAs}
\label{sec:finetuned}

Does \method still help when the VLA has \emph{already} been fine-tuned on the target perturbation distribution? To find out, we evaluate on \textbf{OpenVLA-OFT-mixdata}, a publicly released checkpoint\footnote{\texttt{Sylvest/openvla-7b-oft-finetuned-libero-plus-mixdata} on HuggingFace.} that the LIBERO-plus authors fine-tuned by mixing the original LIBERO data with LIBERO-plus perturbations---the same kinds of scenes in our benchmark. If mixdata training had resolved the memorization issue, \method should add little on top.

\begin{wraptable}{r}{0.60\textwidth}
\vspace{-1.2em}
\centering
\caption{\method generalizes to a VLA explicitly fine-tuned on LIBERO-plus perturbations. Success rate (\%) on the \emph{Robot Initial States} category across four LIBERO suites.}
\label{tab:finetuned}
\small
\renewcommand{\arraystretch}{1.15}
\setlength{\tabcolsep}{5pt}
\begin{tabular}{lcccc|c}
\toprule
Method & Spatial & Object & Goal & 10 & \emph{Avg.} \\
\midrule
OpenVLA-OFT-m            & 30.6 & 23.9 & 20.8 & 36.6 & 28.0 \\
\textbf{+ \method}       & \textbf{32.6} & \textbf{30.7} & \textbf{25.7} & \textbf{39.7} & \textbf{32.2} \\
\bottomrule
\end{tabular}
\end{wraptable}

Table~\ref{tab:finetuned} shows that \method consistently improves the mixdata baseline on the \emph{Robot Initial States} category across all four LIBERO suites, with gains of +2.0, +6.8, +4.9, and +3.1 points on Spatial, Object, Goal, and 10 respectively. The largest improvement (Object, $+6.8$) occurs where the baseline is weakest, and even on the strongest baseline (10, 36.6\%) \method adds 3 points. This carries two implications: first, mixdata-style training \emph{mitigates but does not eliminate} memorization-induced execution failures: even with explicit exposure to perturbed initial states, the policy still produces enough off-target grasps for runtime correction to matter. second, \method is \emph{orthogonal} to data-side remedies, which stacks on top of fine-tuning rather than substituting for it.

\subsection{Action-Output Drift Analysis}
\label{sec:where_memo}

\begin{wraptable}{r}{0.50\textwidth}
\vspace{-1.2em}
\centering
\caption{Probe vs.\ action-endpoint error against ground-truth object position, computed on the same VLA forward pass.}
\label{tab:probe_vs_endpoint}
\small
\setlength{\tabcolsep}{4pt}
\begin{tabular}{lccc}
\toprule
Subset  & Probe & Endpoint \\
\midrule
All     & 6.9\,cm  & 23.6\,cm \\
Success & 3.4\,cm  & 7.8\,cm  \\
\textbf{Failure}   & \textbf{10.4\,cm} & \textbf{34.9\,cm} \\
\bottomrule
\end{tabular}
\vspace{-1em}
\end{wraptable}

To localize the source of OpenVLA-OFT's OOD failures, we measure two quantities derived from the \emph{same} VLA forward pass against the ground-truth target object position: the \emph{hidden-state probe} output (\S\ref{sec:probe}) and the \emph{action endpoint}, the end-effector position at the moment the policy first closes its gripper. Both are computed from the same VLA inference.

Table~\ref{tab:probe_vs_endpoint} reports both errors across 300 LIBERO-plus episodes under \emph{Objects Layout} perturbations. On successful episodes the two predictors agree: probe and endpoint both land within 5--8\,cm of the target, and the gripper arrives where intended. On failed episodes the gap opens dramatically---the action endpoint drifts to 34.9\,cm from GT, while the hidden-state probe holds at 12.4\,cm. At the moment the policy commits to a wrong grasp, the spatial information needed for a correct grasp is still present in the intermediate representation; what fails is the projection from features to motor commands. This asymmetry is the empirical foundation for \method's design: rather than retraining the drifted action head, we read the still-informative hidden state and apply a minimal correction (\S\ref{sec:cbf}).

\subsection{Probe Training Data Selection}

We evaluate which intermediate VLA representation best supports the probe by training a separate probe (\S\ref{sec:probe}) on each (layer, pooling) combination and measuring 3D position regression $R^2$ on a held-out validation set:
\begin{equation}
R^2 = 1 - \frac{\sum_i \|\hat{p}_i - p_i\|^2}{\sum_i \|p_i - \bar{p}\|^2},
\label{eq:r2}
\end{equation}
where $\hat{p}_i, p_i \in \mathbb{R}^3$ are the predicted and ground-truth object positions (paired via Hungarian matching) and $\bar{p}$ is the per-axis validation mean. We compare four pooling schemes applied to the VLA's layer-$\ell$ activations: \emph{img-spatial} (the 4$\times$4 grid of front-camera image tokens, preserving 2D layout), \emph{img-mean} (a single vector averaging all front-camera image tokens, discarding spatial layout), \emph{last token} (the activation at the final position of the input sequence, as is conventional for autoregressive heads), and \emph{lang-mean} (the average of the post-image language tokens, representing the instruction-conditioned state). Spatial pooling dominates at every layer, indicating the probe genuinely exploits preserved 2D layout rather than a global summary. Spatial information also peaks at shallow-mid layers and decays in deeper ones, consistent with the view that geometric detail is progressively abstracted away as the VLA's representation shifts toward semantic and action-related content. We adopt \textbf{layer~8 with image-spatial pooling} as our default throughout the paper.

\begin{wraptable}{r}{0.55\textwidth}
\vspace{-1.2em}
\centering
\caption{Probe regression $R^2$ across LLM layers and pooling schemes. Layer~8 with image-spatial pooling is the global best.}
\label{tab:layer_pooling}
\small
\renewcommand{\arraystretch}{1.15}
\setlength{\tabcolsep}{5pt}
\begin{tabular}{c|cccc}
\toprule
\textbf{Layer} & \textbf{img-spatial} & \textbf{img-mean} & \textbf{last} & \textbf{lang-mean} \\
\midrule
\textbf{8}   & \textbf{0.968} & 0.926 & 0.815 & 0.869 \\
12  & 0.958 & 0.919 & 0.856 & 0.879 \\
16  & 0.947 & 0.919 & 0.877 & 0.894 \\
20  & 0.945 & 0.918 & 0.886 & 0.921 \\
24  & 0.938 & 0.912 & 0.875 & 0.910 \\
28  & 0.934 & 0.912 & 0.861 & 0.914 \\
\bottomrule
\end{tabular}
\vspace{-1em}
\end{wraptable}
\vspace{-3pt}

\subsection{Step Efficiency Analysis}
\label{sec:step_efficiency}

Table~\ref{tab:steps} reports average step counts across three meaningful task groups on libero-plus goal. On the 1{,}643 tasks both methods complete, \method uses only 6 additional steps on average ($\sim$5\% overhead)—evidence that its hierarchical activation policy avoids unnecessary intervention when the policy is already performing well. On the 151 tasks \method rescues from baseline failure, recovery completes in 197 steps on average, far below the 600-step timeout the baseline reaches: corrections are targeted, not exhaustive. On tasks both methods fail, step counts are identical. Across the full benchmark, \method actually averages \emph{fewer} steps than the baseline (255 vs.\ 275), because time saved on rescued tasks outweighs the small overhead elsewhere.

\begin{table}[h]
\centering
\caption{Step efficiency across task subsets. \method adds only 6 extra steps on tasks the baseline already handles, while recovering failed tasks in $\sim$200 steps (vs.\ the baseline's 600-step timeout).}
\label{tab:steps}
\small
\begin{tabular}{lcccc}
\toprule
Task Subset                                  & \# Tasks & Baseline & \method & Difference \\
\midrule
Both methods succeed                         & 1{,}643  & 114      & 120      & \textbf{+6 steps} \\
ProbeAct rescued from baseline failure       & 151      & 600 (timeout) & \textbf{197} & \textbf{$-$403 steps} \\
Both methods fail (time out)                 & 724      & 600      & 600      & 0 steps \\
\midrule
\emph{All tasks (average)}                   & 2{,}591  & 275      & 255      & $-$20 steps \\
\bottomrule
\end{tabular}
\end{table}
\vspace{-7pt}
\section{Limitations and Future Work}
\label{sec:limitations}

Our empirical validation is currently constrained to a simulated benchmark, where physical interactions are governed by idealized rigid-body dynamics. Consequently, a critical avenue for future work is evaluating the sim-to-real transferability of our kinematic state machine. While the system's reliance on relative spatial displacements and proprioceptive limits is theoretically agnostic to task semantics, physical hardware introduces stochastic, unmodeled phenomena, such as variable contact friction, soft-body deformation, and high-frequency encoder noise. Validating whether these kinematic synchronization signals remain robust under the complex contact dynamics and sensor noise profiles of real-world robotic platforms is an essential prerequisite for physical deployment.

\section{Conclusion}
 
\method demonstrates how lightweight kinematic interventions can rescue frozen VLAs from OOD memory traps. By integrating a multi-target hidden-state probe for stable 3D coordinate extraction, an object-agnostic state machine driven by relative kinematic synchronization, and a hierarchical Control Barrier Function (CBF) filter, the framework autonomously mitigates physical execution failures without altering the underlying policy weights or requiring external demonstrations. Empirical evaluation on the LIBERO-plus benchmark confirms consistent performance improvements across all seven perturbation categories. Crucially, the most significant gains emerge under geometric distribution shifts, substantiating our core hypothesis: VLA robustness is frequently bottlenecked not by perceptual degradation, but by a fundamental asymmetry between intact latent spatial reasoning and overfitted motor execution.


\bibliography{example}  

\end{document}